\documentclass[conference]{IEEEtran}

\usepackage{filecontents,amsmath,graphicx}
\usepackage[square,numbers]{natbib}
\ifCLASSOPTIONcompsoc
\usepackage[caption=false,font=normalsize,labelfon
t=sf,textfont=sf]{subfig}
\else
\usepackage[caption=false,font=footnotesize]{subfi
	g}
\fi

\IEEEoverridecommandlockouts

\title{An Investigation on Deep Learning with Beta Stabilizer}

\author{\IEEEauthorblockN{Qi Liu, Tian Tan, Kai Yu}
\IEEEauthorblockA{Key Laboratory of Shanghai Education Commission for Intelligent Interaction and Cognitive Engineering \\
	SpeechLab, Department of Computer Science and Engineering \\
	Brain Science and  Technology Research Center \\
	Shanghai Jiao Tong University, Shanghai, China\\
Emails: \{liuq901, tantian, kai.yu\}@sjtu.edu.cn}
}

\begin{document}

\maketitle

\begin{abstract}
Artificial neural networks (ANN) have been used in many applications such like handwriting recognition and speech recognition. It is well-known that learning rate is a crucial value in the training procedure for artificial neural networks. It is shown that the initial value of learning rate can confoundedly affect the final result and this value is always set manually in practice. A new parameter called beta stabilizer has been introduced to reduce the sensitivity of the initial learning rate. But this method has only been proposed for deep neural network (DNN) with sigmoid activation function. In this paper we extended beta stabilizer to long short-term memory (LSTM) and investigated the effects of beta stabilizer parameters on different models, including LSTM and DNN with relu activation function. It is concluded that beta stabilizer parameters can reduce the sensitivity of learning rate with almost the same performance on DNN with relu activation function and LSTM. However, it is shown that the effects of beta stabilizer on DNN with relu activation function and LSTM are fewer than the effects on DNN with sigmoid activation function.
\end{abstract}

\section{Introduction}
Hidden Markov model (HMM) \cite{baum1966} and Gaussian mixture model (GMM) \cite{spall1992} have been used to solve handwriting recognition and speech recognition problem for a long time \cite{gales2008}.
Due to the limitation of GMM, ANN especially DNN \cite{hinton2006} and recurrent neural network (RNN) \cite{goller1996} have been used to combined with HMM and provides huge improvement on the performance \cite{hinton2012}. 

The state-of-art training method for ANN nowadays is mini-batch based stochastic gradient descent (SGD) with momentum \cite{lecun1998}. For SGD algorithm, learning rate is a crucial and sensitive value. The initial value of learning rate has huge effects on the final performance and converge speed for ANN training. However, this value is always an experience parameter i.e. which is set manually. Another problem is that the best initial value of learning rate can be varied with different tasks, different neural network structures and different toolkits. How to set the initial value of learning rate is a tricky part in the training procedure of ANN.

Some researchers are using grid search on learning to choose the best initial value \cite{bergstra2012}. Some provides self adjustment techniques in pre-training to automatically select the initial value \cite{george2011}. There are also many training algorithms, which are not such sensitive with learning rate, are produced to solve this problem. These methods include AdaDelta \cite{zeiler2012}, AdaGrad \cite{duchi2010} and natural gradient \cite{amari1998}.

\cite{ghahremani2016} provide a quite different solution. For every linear transform parameter, a learnable scalar parameter is added. This parameter can affect the update procedure in SGD with learning rate together. By combining the original learning rate and this parameter, the learning rate can be learnable. This can reduce the sensitivity of initial learning rate and accelerate the converge speed.

In \cite{ghahremani2016}, only DNN with sigmoid activation function has been used. However, DNN with relu activation functions converges quickly than DNN with sigmoid function, and has been successfully applied to many applications \cite{nair2010} \cite{maas2013}. LSTM has been the state-of-art solution for speech recognition and handwriting recognition \cite{gers2003} \cite{graves2013} \cite{graves2014} because it has the ability to model sequential data. End-to-end models including connectionist temporal classification (CTC) \cite{graves2006} \cite{liu2015} and attention model \cite{chan2016} \cite{chorowski2015} also use LSTM widely. Therefore it is significant to evaluate beta stabilizer on these new neural network models. In this paper, we extend beta stabilizer parameters to LSTM and evaluate the effects on different ANN architectures including LSTM and DNN with relu activation function.

Multiple speech recognition experiments have been done to verify the results. Two data corpora are prepared, one is the local 15 hours Chinese dataset, the other one is Switchboard 50 hours English dataset \cite{godfrey1992}. The neural network structure contains DNN with sigmoid function, DNN with relu function and deep LSTM. All the experiments using the same SGD algorithm on a single CUDA based GPU.

The experimental results show that beta stabilizer parameters achieve good results in DNN with sigmoid function. In some cases the performance will be reduced in DNN with relu function and LSTM. However, the sensitivity of initial learning rate can always be reduced with beta stabilizer parameters regardless of neural network architectures.

The rest of the paper is organized as follows: Section II gives the detail of beta stabilizer for DNN. We show how to extend beta stabilizer to LSTM in section III. Section IV shows the setup and results of our experiments. Finally, the conclusion can be found in section V and discussion can be found in section VI.

\section{Overview of Beta Stabilizer}

\subsection{SGD Background}
SGD is a first order optimization algorithm. It is based on a differentiable function decrease fastest along the negative direction of its gradient. This algorithm is well used in machine learning field to optimize a model with multiple variables.

To use SGD in the training procedure of ANN, the parameter $\theta$ should be updated to minimize the objective loss function $\mathcal{L}$. For every scale value $\theta_i$ in $\theta$, the gradient $$\Delta_i=\frac{\partial\mathcal{L}}{\partial\theta_i}$$ will be calculated. After all the gradients has been calculated, the update procedure $$\theta_i=\theta_i-\eta\Delta_i$$ will be applied. Here, $\eta$ is the learning rate.

\subsection{Learning Rate Scheduling Method}
In practice, the learning rate may be changed during the training procedure. The process of the adjustment of learning rate is called learning rate scheduling. There are several methods for learning rate scheduling. Two widely used methods are early stopping \cite{yao2007} and learning rate halving \cite{goodfellow2016}. Early stopping will terminate the training procedure when performance on cross validation set consecutively becomes worse in some iterations. Learning rate halving will reduce the learning rate by half when the performance on cross validation set becomes worse. 

There are also some new techniques on learning rate scheduling such like exponential scheduling \cite{bottou2010}, learning rate monitor \cite{schaul2012} and learning rate auto-adjustment method \cite{yu2014}.

Due to the sensitivity of initial learning rate also relies on the learning rate scheduling method. To control the experimental variables, we use learning rate halving as our learning rate scheduling method in all the experiments.

\subsection{Beta Stabilizer for DNN}
The beta stabilizer parameter is a scalar parameter for each layer in DNN. For normal DNN hidden layers, the formula is $$\mathbf{y}=\mathbf{Wx+b}$$
here $\mathbf{x}$ is the input vector and $\mathbf{y}$ is the output vector. $\mathbf{W}$ is the linear transform parameter matrix and $\mathbf{b}$ is the bias parameter vector.

With a scalar beta stabilizer parameter, the formula change to $$\mathbf{y}=e^\beta\mathbf{Wx+b}$$ where $e$ is the base of natural logarithm and $\beta$ is the stabilizer parameter.

In the training procedure, the propagation phase can be done by directly following the above formula. The back-propagation phase need to calculated the gradient of objective function respect to $\mathbf{x}$, $\mathbf{W}$, $\mathbf{b}$ and $\beta$.

The gradients respect to $\mathbf{x}$ and $\mathbf{W}$ have minor changes, $$\frac{\partial\mathcal{L}}{\partial\mathbf{x}}= e^\beta\mathbf{W}^T\frac{\partial\mathcal{L}}{\partial\mathbf{y}}$$ and 
$$\frac{\partial\mathcal{L}}{\partial\mathbf{W}}= e^\beta\frac{\partial\mathcal{L}}{\partial\mathbf{y}}\mathbf{x}^T.$$

The gradient respect to $\mathbf{b}$ remains unchanged, $$\frac{\partial\mathcal{L}}{\partial\mathbf{b}}= \frac{\partial\mathcal{L}}{\partial\mathbf{y}}.$$

The final problem is how to update the stabilizer parameter $\beta$. By the chain rule, $$\frac{\partial\mathcal{L}}{\partial\beta} =\frac{\partial\mathcal{L}}{\partial\mathbf{y}} \frac{\partial\mathbf{y}}{\partial\beta}= e^\beta\frac{\partial\mathcal{L}}{\partial\mathbf{y}}^T\mathbf{Wx}.$$
Due to $$\frac{\partial\mathcal{L}}{\partial\mathbf{x}}^T= e^\beta\frac{\partial\mathcal{L}}{\partial\mathbf{y}}^T\mathbf{W},$$
we have $$\frac{\partial\mathcal{L}}{\partial\beta}= \frac{\partial\mathcal{L}}{\partial\mathbf{x}}^T\mathbf{x}$$
i.e. the inner product of $\frac{\partial\mathcal{L}}{\partial\mathbf{x}}$ and $\mathbf{x}$. The update rule is $$\beta=\beta-\eta\frac{\partial\mathcal{L}}{\partial\mathbf{x}}^T\mathbf{x}.$$
This means the value of $\beta$ relies on the relation between layer input and its gradient. It shows that $\beta$ will be increased if scaling $\mathbf{x}$ up can improve the performance and vice versa.

It is shown that $\beta$ relies on the value and gradient of input vector $\mathbf{x}$. For DNN with multiple hidden layers, these values will depend on the activation function. This is the reason why we investigate the performance of beta stabilizer in DNN with relu activation function.

At the beginning of training procedure, all $\beta$ values are set to 0 thus $e^\beta=1$ where the initial model remains same with the one without stabilizer parameter.

\section{Beta Stabilizer for LSTM}
LSTM is an architecture that uses memory cell to keep information \cite{hochreiter1997}, and becomes the state-of-art solution for speech recognition and handwriting recognition nowadays. It can be implemented by the following formulas: 
\begin{align*}
\mathbf{i_t}&=\mathbf{\sigma(W_{xi}x_t+W_{hi}h_{t-1}+W_{ci}c_{t-1}+b_i)} \\
\mathbf{f_t}&=\mathbf{\sigma(W_{xf}x_t+W_{hf}h_{t-1}+W_{cf}c_{t-1}+b_f)} \\
\mathbf{c_t}&=\mathbf{f_t\cdot c_{t-1}+i_t\cdot\tanh(W_{xc}x_t+W_{hc}h_{t-1}+b_c)} \\
\mathbf{o_t}&=\mathbf{\sigma(W_{xo}x_t+W_{ho}h_{t-1}+W_{co}c_t+b_o)} \\
\mathbf{h_t}&=\mathbf{o_t\cdot\tanh(c_t)}
\end{align*}
here $\sigma$ is sigmoid function.

In DNN, beta stabilizer is applied to the linear transform matrix. But in one LSTM layer, there three gates and one main affine operation. Three ways have been considered to extend beta stabilizer to LSTM. Layer shared beta stabilizer, gate shared beta stabilizer and independent beta stabilizer.

Layer shared beta stabilizer means a single $e^\beta$ will be added for all the linear transform operation in one LSTM layer. Gate shared beta stabilizer means every individual gate in LSTM will have a beta stabilizer. Because we believe that beta stabilizer is a kind of normalization of linear transform matrix. We have calculated the l2-norm of every linear transform matrix of a trained LSTM model. It is found that the l2-norm of matrices of cell values (i.e. $\mathbf{W_{ci}, W_{cf}, W_{co}}$) is one magnitude less than other matrices. The l2-norm of matrices of input vector (i.e. $\mathbf{W_{xi}, W_{xf}, W_{xc},  W_{xo}}$) in the first LSTM layer is half of the l2-norm of matrices of hidden activations (i.e. $\mathbf{W_{hi}, W_{hf}, W_{hc},  W_{ho}}$). This shows shared beta stabilizer may be not suitable for LSTM.

Therefore, independent beta stabilizer has been selected as our solution. For every linear transform operation, a beta stabilizer has been added. We believed that independent beta stabilizer can adjust the scale of every matrix separately and appropriately. The changed formulas are shown below:
\begin{align*}
\mathbf{i_t}&=\sigma(e^{\beta_{xi}}\mathbf{W_{xi}x_t}+ e^{\beta_{hi}}\mathbf{W_{hi}h_{t-1}}+ e^{\beta_{ci}}\mathbf{W_{ci}c_{t-1}+b_i)} \\
\mathbf{f_t}&=\sigma(e^{\beta_{xf}}\mathbf{W_{xf}x_t}+ e^{\beta_{hf}}\mathbf{W_{hf}h_{t-1}}+ e^{\beta_{cf}}\mathbf{W_{cf}c_{t-1}+b_f)} \\
\mathbf{c_t}&=\mathbf{f_t\cdot c_{t-1}+i_t\cdot}
\tanh(e^{\beta_{xc}}\mathbf{W_{xc}x_t}+
e^{\beta_{hc}}\mathbf{W_{hc}h_{t-1}+b_c)} \\
\mathbf{o_t}&=\sigma(e^{\beta_{xo}}\mathbf{W_{xo}x_t}+ e^{\beta_{ho}}\mathbf{W_{ho}h_{t-1}}+e^{\beta_{co}}\mathbf{W_{co}c_t+b_o)} \\
\mathbf{h_t}&=\mathbf{o_t\cdot\tanh(c_t)}
\end{align*}
The back-propagation and update rule can be derived by using the similar methods in section II. 

\section{Experiments}

\begin{table}[!t]
	\caption{Performance of local Chinese dataset with sigmoid DNN.}
	\label{chn-sigmoid}
	\centering
	\begin{tabular}{|c|c|c|c|c|}
		\hline
		Init LR & With stabilizer & CE on CV & Frame ACC on CV & WER \\
		\hline
		0.8 & False & 2.842 & 40.6\% & 31.14\% \\
		\hline
		0.1 & False & 3.394 & 33.7\% & 45.98\% \\
		\hline
		0.8 & True & 2.835 & 42.4\% & 29.49\% \\
		\hline
		0.1 & True & 2.830 & 41.9\% & 30.54\% \\
		\hline
		0.01 & True & 2.772 & 41.6\% & 30.80\% \\
		\hline
	\end{tabular}
\end{table}

\begin{table}[!t]
	\caption{Performance of local Chinese dataset with relu DNN.}
	\label{chn-relu}
	\centering
	\begin{tabular}{|c|c|c|c|c|}
		\hline
		Init LR & With stabilizer & CE on CV & Frame ACC on CV & WER \\
		\hline
		0.8 & False & 3.149 & 40.0\% & 32.68\% \\
		\hline
		0.1 & False & 2.796 & 43.4\% & 28.97\% \\
		\hline
		0.0125 & False & 2.622 & 43.4\% & 29.19\% \\
		\hline
		0.0016 & False & 3.022 & 38.0\% & 39.43\% \\
		\hline		
		0.1 & True & 2.962 & 42.1\% & 30.74\% \\
		\hline
		0.0125 & True & 2.802 & 41.7\% & 29.87\% \\
		\hline
		0.0016 & True & 2.907 & 39.9\% & 32.53\% \\
		\hline		
	\end{tabular}
\end{table}

\begin{table}[!t]
	\caption{Performance of local Chinese dataset with LSTM.}
	\label{chn-lstm}
	\centering
	\begin{tabular}{|c|c|c|c|c|}
		\hline
		Init LR & With stabilizer & CE on CV & Frame ACC on CV & WER \\
		\hline
		0.1 & False & 2.099 & 51.3\% & 26.04\% \\
		\hline
		0.04 & False & 2.057 & 51.4\% & 26.25\% \\
		\hline
		0.005 & False & 2.113 & 50.2\% & 26.88\% \\
		\hline
		0.0006 & False & 3.236 & 36.6\% & 47.01\% \\
		\hline		
		0.1 & True & 2.133 & 50.0\% & 25.86\% \\
		\hline
		0.04 & True & 2.171 & 49.8\% & 26.68\% \\
		\hline
		0.005 & True & 2.228 & 49.7\% & 27.04\% \\
		\hline
		0.0006 & True & 2.366 & 47.0\% & 29.35\% \\
		\hline				
	\end{tabular}
\end{table}

\begin{table}[!t]
	\caption{Performance of Switchboard English dataset with sigmoid DNN.}
	\label{swbd-sigmoid}
	\centering
	\begin{tabular}{|c|c|c|c|c|}
		\hline
		Init LR & With stabilizer & CE on CV & Frame ACC on CV & WER \\
		\hline
		0.8 & False & 1.991 & 50.6\% & 21.3\% \\
		\hline
		0.1 & False & 2.270 & 44.8\% & 27.1\% \\
		\hline
		0.8 & True & 2.235 & 48.6\% & 21.4\% \\
		\hline
		0.1 & True & 2.197 & 48.1\% & 22.4\% \\
		\hline
		0.01 & True & 2.133 & 48.1\% & 23.5\% \\
		\hline
	\end{tabular}
\end{table}

\begin{table}[!t]
	\caption{Performance of Switchboard English dataset with relu DNN.}
	\label{swbd-relu}
	\centering
	\begin{tabular}{|c|c|c|c|c|}
		\hline
		Init LR & With stabilizer & CE on CV & Frame ACC on CV & WER \\
		\hline
		0.8 & False & 2.378 & 46.9\% & 23.4\% \\
		\hline
		0.1 & False & 2.371 & 47.3\% & 22.4\% \\
		\hline
		0.0125 & False & 2.099 & 48.5\% & 23.3\% \\
		\hline
		0.0016 & False & 2.267 & 45.4\% & 27.0\% \\
		\hline
		0.1 & True & 2.263 & 48.0\% & 22.4\% \\
		\hline
		0.0125 & True & 2.164 & 47.7\% & 23.2\% \\
		\hline
		0.0016 & True & 2.245 & 45.8\% & 26.2\% \\
		\hline		
	\end{tabular}
\end{table}

\begin{table}[!t]
	\caption{Performance of Switchboard English dataset with LSTM.}
	\label{swbd-lstm}
	\centering
	\begin{tabular}{|c|c|c|c|c|}
		\hline
		Init LR & With stabilizer & CE on CV & Frame ACC on CV & WER \\
		\hline
		0.1 & False & 1.582 & 59.2\% & 20.5\% \\
		\hline
		0.04 & False & 1.561 & 59.2\% & 20.5\% \\
		\hline
		0.005 & False & 1.616 & 57.8\% & 21.7\% \\
		\hline
		0.0006 & False & 1.807 & 54.0\% & 25.7\% \\
		\hline		
		0.1 & True & 1.649 & 57.8\% & 21.2\% \\
		\hline
		0.04 & True & 1.628 & 58.3\% & 20.8\% \\
		\hline
		0.005 & True & 1.665 & 57.3\% & 22.2\% \\
		\hline
		0.0006 & True & 1.774 & 55.1\% & 23.2\% \\
		\hline					
	\end{tabular}
\end{table}

\subsection{Experimental Setup}
Two speech recognition corpora are used in our experiments. The first one is local 15 hours Chinese dataset. The second corpus is Switchboard 50 hours English dataset.

For every dataset, three network structures are prepared. These including DNN with sigmoid function, DNN with relu function and LSTM.

For the experiments with the same corpus and structures, only the initial learning rate may be varied, all the other parameters are the same. Learning rate halving has been used as the learning rate scheduling method. All the experiments are done on a single CUDA based GPU card.

\begin{figure*}[!t]
	\centering
	\subfloat[Sigmoid DNN]{\includegraphics[width=3.3in]{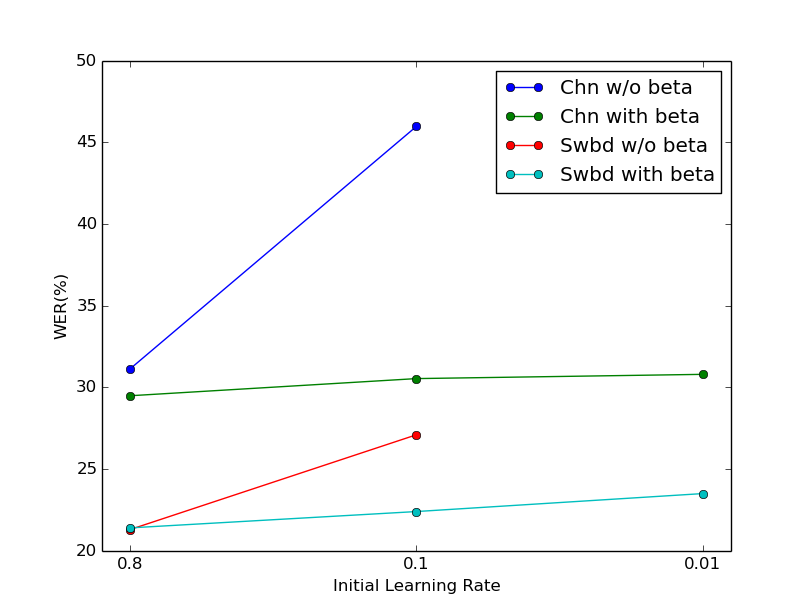}
		\label{fig_sigmoid}}
	\hfil
	\subfloat[Relu DNN]{\includegraphics[width=3.3in]{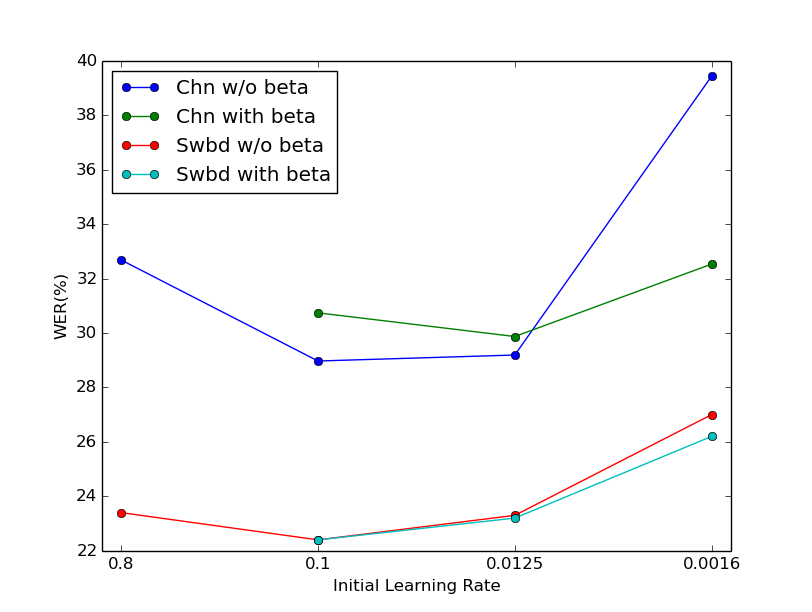}
		\label{fig_relu}}
	\hfil
	\subfloat[LSTM]{\includegraphics[width=3.3in]{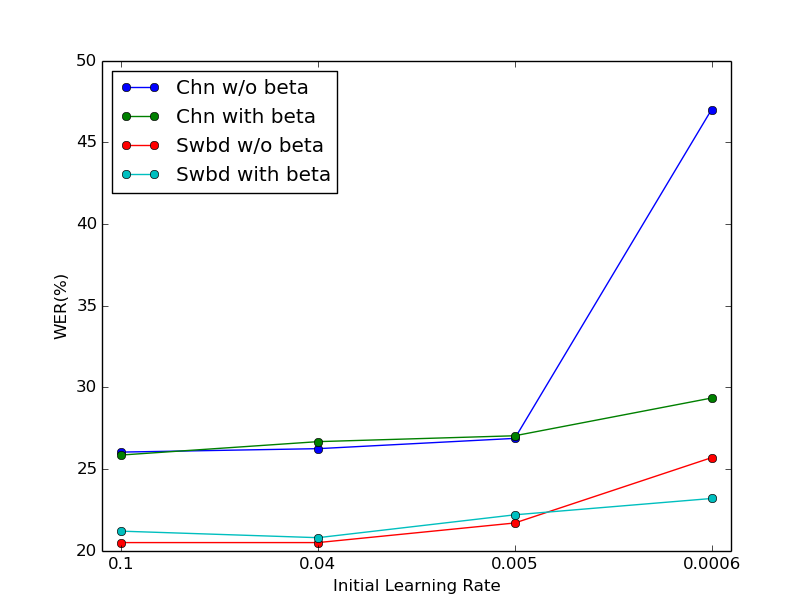}
		\label{fig_lstm}}	
	\caption{Experimental results of different ANN architectures. Chn stands for local Chinese dataset and Swbd stands for Switchboard English dataset in the legends.}
	\label{fig}
\end{figure*}

\subsection{Experimental Results for Local Chinese Dataset}
The local Chinese dataset contains 15 hours data for the training set. For this dataset, we use the DNN model with 6 hidden layers contains 1024 nodes and LSTM model with 3 hidden layers contains 600 nodes.

Table \ref{chn-sigmoid} shows the results on the DNN with sigmoid activation function. It is clear that without beta stabilizer, changing the initial learning rate from 0.8 to 0.1 has huge impact on the final performance. While the initial learning rate has almost no effect on the performance with beta stabilizer.

Table \ref{chn-relu} concludes the performance on the DNN with relu activation function. It shows that beta stabilizer did not work as well as in sigmoid DNN. The best WER with beta stabilizer becomes a little worse than without it. However, is also claims that the results with beta stabilizer are less sensitive than without it.

Table \ref{chn-lstm} is the performance on LSTM. The results are almost the same when the initial learning rate is suitable. But when the initial value becomes relative small, the network with beta stabilizer can achieve better performance.

\subsection{Experimental Results for Switchboard English Dataset}
For Switchboard 50 hours dataset, we use DNN with 6 hidden layers contains 2048 nodes and LSTM with 3 hidden layers contais 1024 nodes.

Table \ref{swbd-sigmoid}, \ref{swbd-relu} and \ref{swbd-lstm} shows the results of sigmoid DNN, relu DNN and LSTM respectively. The performance of sigmoid DNN has been improved. But compared with the results of local Chinese dataset, the performance of relu DNN becomes better while the performance of LSTM be slightly worse. However, it can be concluded  that the networks with beta stabilizer is less sensitive than the networks without it from these tables.

\section{Conclusion}

From the above results, we conclude that beta stabilizer parameters can reduce the sensitivity of results respect to initial learning rate in both DNN and LSTM. Figure \ref{fig} clear shows the results. From figure \ref{fig_sigmoid}, it is clear that beta stabilizer parameters achieve the best performance on DNN with sigmoid function. From figure \ref{fig_relu} and \ref{fig_lstm}, DNN with relu function and LSTM are less sensitive about the initial learning rate than DNN with sigmoid function. However, when the initial learning rate becomes relative small, i.e. 0.0016 for relu DNN and 0.0006 for LSTM, the networks with beta stabilizer parameters still give acceptable results. Even with extremely small initial value (0.0001) of learning rate, LSTM with beta stabilizer still can give reasonable results while LSTM without it cannot converge at all. 

It is observed that the effects of beta stabilizer on DNN with relu function and LSTM are fewer than the effects on DNN with sigmoid function. The performance may be worse with suitable initial learning rate when beta stabilizer has been used for DNN with relu function and LSTM. However, beta stabilizer performs well when the initial value is relative small. We concluded that beta stabilizer could reduce the sensitivity of initial learning rate with multiple ANN architectures.

\section{Discussion}

In some complicated network such like convolution-LSTM-deep neural network (CLDNN) \cite{sainath2015} \cite{simard2003} and multi-task network \cite{yin2016}, different parts of the network can have different beta stabilizers. \cite{ghahremani2016} also mentioned that beta stabilizer can be used not only for SGD but also other training algorithm such like AdaGrad and AdaDelta.

Therefore our ongoing and future works include 1) observe the results of beta stabilizer parameters on large scale data, 2) try beta stabilizer parameters with other training algorithms, 3) add beta stabilizer parameters to complicated networks.

\section*{Acknowledgement}
This work was supported by the Shanghai Sailing Program No. 16YF1405300, the Program for Professor of Special Appointment (Eastern Scholar) at Shanghai Institutions of Higher Learning, the China NSFC projects (No. 61573241 and No. 61603252) and the Interdisciplinary Program (14JCZ03) of Shanghai Jiao Tong University in China.

\renewcommand{\bibfont}{\footnotesize}
\bibliographystyle{ieeetr}
\bibliography{exp_beta_paper.bib}

\begin{thebibliography}{10}

\bibitem{baum1966}
L.~E. Baum and T.~Petrie, ``Statistical inference for probabilistic functions
  of finite state markov chains,'' {\em The Annals of Mathematical Statistics},
  vol.~37, no.~6, pp.~1554--1563, 1966.

\bibitem{spall1992}
J.~C. Spall and J.~L. Maryak, ``A feasible bayesian estimator of quantiles for
  projectile accuracy from non-i.i.d. data,'' {\em Journal of the American
  Statistical Association}, vol.~87, no.~419, pp.~676--681, 1992.

\bibitem{gales2008}
M.~Gales and S.~Young, ``The application of hidden markov models in speech
  recognition,'' {\em Foundations and trends in signal processing}, vol.~1,
  no.~3, pp.~195--304, 2007.

\bibitem{hinton2006}
G.~E. Hinton, S.~Osindero, and Y.~W. Teh, ``A fast learning algorithm for deep
  belief nets,'' {\em Neural Computation}, vol.~18, no.~7, p.~1527–1554,
  2006.

\bibitem{goller1996}
C.~Goller and A.~K{\"ü}chler, ``Learning task-dependent distributed
  representations by backpropagation through structure,'' {\em IEEE
  Transactions on Neural Networks}, vol.~1, pp.~347--352, 1996.

\bibitem{hinton2012}
G.~E. Hinton, L.~Deng, D.~Yu, G.~Dahl, A.-r. Mohamed, N.~Jaitly, A.~Senior,
  V.~V., P.~Nguyen, T.~Sainath, and B.~Kingsbury, ``Deep neural networks for
  acoustic modeling in speech recognition: The shared views of four research
  groups,'' {\em Signal Processing Magazine}, vol.~29, no.~6, pp.~82--97, 2012.

\bibitem{lecun1998}
Y.~LeCun, L.~Bottou, G.~B. Orr, and K.-R. M{\"u}ller, ``Efficient backprop,''
  in {\em Neural Networks: Tricks of the Trade}, pp.~9--50, Springer Berlin
  Heidelberg, 1998.

\bibitem{bergstra2012}
J.~Bergstra and Y.~Bengio, ``Random search for hyper-parameter optimization,''
  {\em The Journal of Machine Learning Research}, vol.~13, no.~1, pp.~281--305,
  2012.

\bibitem{george2011}
A.~P. George and W.~B. Powell, ``Adaptive stepsizes for recursive estimation
  with applications in approximate dynamic programming,'' {\em Machine
  Learning}, vol.~65, no.~1, pp.~167--198, 2011.

\bibitem{zeiler2012}
M.~D. Zeiler, ``{ADADELTA:} an adaptive learning rate method,'' 2012.
\newblock arXiv:1212.5701.

\bibitem{duchi2010}
J.~Duchi, E.~Hazan, and Y.~Singer, ``Adaptive subgradient methods for online
  learning and stochastic optimization,'' {\em The Journal of Machine Learning
  Research}, vol.~12, pp.~2121--2159, 2010.

\bibitem{amari1998}
S.-I. Amari, ``Natural gradient works efficiently in learning,'' {\em Neural
  Computation}, vol.~10, no.~2, pp.~251--276, 1998.

\bibitem{ghahremani2016}
P.~Ghahremani and J.~Droppo, ``Self-stabilized deep neural network,'' in {\em
  IEEE International Conference on Acoustics, Speech and Signal Processing},
  pp.~5450--5454, 2016.

\bibitem{nair2010}
V.~Nair and G.~E. Hinton, ``Rectified linear units improve restricted boltzmann
  machines,'' in {\em Proceedings of the International Conference on Machine
  Learning}, pp.~807--814, 2010.

\bibitem{maas2013}
A.~L. Maas, A.~Y. Hannun, and A.~Y. Ng, ``Rectifier nonlinearities improve
  neural network acoustic models,'' in {\em Proceedings of the International
  Conference on Machine Learning}, vol.~30, 2013.

\bibitem{gers2003}
F.~A. Gers, N.~N. Schraudolph, and J.~Schmidhuber, ``Learning precise timing
  with {LSTM} recurrent networks,'' {\em The Journal of Machine Learning
  Research}, vol.~3, pp.~115--143, 2003.

\bibitem{graves2013}
A.~Graves, M.~A.-r., and G.~Hinton, ``Speech recognition with deep recurrent
  neural networks,'' in {\em IEEE International Conference on Acoustics, Speech
  and Signal Processing}, pp.~6645--6649, 2013.

\bibitem{graves2014}
A.~Graves and N.~Jaitly, ``Towards end-to-end speech recognition with recurrent
  neural networks,'' in {\em Proceedings of the International Conference on
  Machine Learning}, pp.~1764--1772, 2014.

\bibitem{graves2006}
A.~Graves, S.~Fern{\'a}ndez, and F.~Gomez, ``Connectionist temporal
  classification: Labelling unsegmented sequence data with recurrent neural
  networks,'' in {\em Proceedings of the International Conference on Machine
  Learning}, pp.~359--376, 2006.

\bibitem{liu2015}
Q.~Liu, L.~Wang, and Q.~Huo, ``A study on effects of implicit and explicit
  language model information for {DBLSTM-CTC} based handwriting recognition,''
  in {\em IEEE International Conference on Document Analysis and Recognition},
  pp.~461--465, 2015.

\bibitem{chan2016}
W.~Chan, N.~Jaitly, Q.~Le, and O.~Vinyals, ``Listen, attend and spell: A neural
  network for large vocabulary conversational speech recognition,'' in {\em
  IEEE International Conference on Acoustics, Speech and Signal Processing},
  pp.~4960--4964, 2016.

\bibitem{chorowski2015}
J.~Chorowski, D.~Bahdanau, D.~Serdyuk, K.~Cho, and Y.~Bengio, ``Attention-based
  models for speech recognition,'' 2015.
\newblock arXiv:1506.07503.

\bibitem{godfrey1992}
J.~J. Godfrey, E.~C. Holliman, and J.~McDaniel, ``{SWITCHBOARD:} telephone
  speech corpus for research and development,'' in {\em IEEE International
  Conference on Acoustics, Speech and Signal Processing}, vol.~1, pp.~517--520,
  1992.

\bibitem{yao2007}
Y.~Yao, L.~Rosasco, and A.~Caponnetto, ``On early stopping in gradient descent
  learning,'' {\em Constructive Approximation}, vol.~26, no.~2, pp.~289--315,
  2007.

\bibitem{goodfellow2016}
I.~Goodfellow, Y.~Bengio, and A.~Courville, ``Deep learning.'' Book in
  preparation for MIT Press, 2016.

\bibitem{bottou2010}
L.~Bottou, ``Large-scale machine learning with stochastic gradient descent,''
  in {\em International Conference on Computational Statistics}, pp.~177--187,
  2010.

\bibitem{schaul2012}
T.~Schaul, S.~Zhang, and Y.~LeCun, ``No more pesky learning rates,'' 2012.
\newblock arXiv:1206.1106.

\bibitem{yu2014}
D.~Yu, A.~Eversole, M.~L. Seltzer, K.~Yao, Z.~Huang, B.~Guenter, O.~Kuchaiev,
  Y.~Zhang, F.~Seide, H.~Wang, J.~Droppo, G.~Zweig, C.~Rossbach, J.~Currey,
  J.~Gao, A.~May, B.~Peng, A.~Stolcke, and M.~Slaney, ``An introduction to
  computational networks and the computational network toolkit,'' 2014.
\newblock MSR-TR-2014-112.

\bibitem{hochreiter1997}
S.~Hochreiter and J.~Schmidhuber, ``Long short-term memory,'' {\em Neural
  Computation}, vol.~9, no.~8, p.~1735–1780, 1997.

\bibitem{sainath2015}
T.~N. Sainath, O.~Vinyals, A.~Senior, and H.~Sak, ``Convolutional, long
  short-term memory, fully connected deep neural networks,'' in {\em IEEE
  International Conference on Acoustics, Speech and Signal Processing}, 2015.

\bibitem{simard2003}
P.~Y. simard, D.~Steinkraus, and J.~C. Platt, ``Best practices for
  convolutional neural networks applied to visual document analysis,'' in {\em
  IEEE International Conference on Document Analysis and Recognition}, vol.~3,
  pp.~958--962, 2003.

\bibitem{yin2016}
M.~Yin, S.~Sivadas, K.~Yu, and B.~Ma, ``Discriminatively trained joint speaker
  and environment representations for adaptation of deep neural network
  acoustic models,'' in {\em IEEE International Conference on Acoustics, Speech
  and Signal Processing}, pp.~5065--5069, 2016.

\end{thebibliography}

\end{document}